\documentclass[10pt,twocolumn,letterpaper]{article} 
\usepackage{avss}
\usepackage{times}

\usepackage{graphicx}
\usepackage{amsmath,amssymb} 
\usepackage{color}
\usepackage{url}
\usepackage{xspace}
\usepackage{algorithmic}
\usepackage{booktabs}
\usepackage{multirow}
\usepackage{rotating}
\usepackage{balance}
\usepackage[pagebackref=false,breaklinks=true,colorlinks,bookmarks=false,urlcolor=blue,linkcolor=blue,citecolor=blue,pdftitle={Combined Learning of Salient Local Descriptors and Distance Metrics for Image Set Face Verification},pdfauthor={Conrad Sanderson, Mehrtash T. Harandi, Yongkang Wong, Brian C. Lovell}]{hyperref}

\graphicspath{{./}{./figures/}}

\def\Vec#1{{\boldsymbol{#1}}}
\def\Mat#1{{\boldsymbol{#1}}}

\avssfinalcopy 

\def\avssPaperID{xxx} 

\ifavssfinal\fi

\begin{document}

\title
  {
  Combined Learning of Salient Local Descriptors and Distance Metrics\\
  for Image Set Face Verification
  }

\author
  {
  {\it Conrad Sanderson, Mehrtash T. Harandi, Yongkang Wong, Brian C. Lovell}\\
  ~\\
  NICTA, PO Box 6020, St Lucia, QLD 4067, Australia\\
  University of Queensland, School of ITEE, QLD 4072, Australia~%
    \textcolor{white}{\thanks
      {%
      {\bf Published in:}
      IEEE International Conference on Advanced Video and Signal-Based Surveillance (AVSS), pp.~294--299, 2012.
      \href{http://dx.doi.org/10.1109/AVSS.2012.23}{\bf http://dx.doi.org/10.1109/AVSS.2012.23} ~
      Copyright \copyright~2012 IEEE. \mbox{Personal} use of this material is permitted.
      However, permission to use this material for any other purposes must
      be obtained from the IEEE by sending an email to
      pubs-permissions@ieee.org%
      }}
  }

\maketitle
\ifavssfinal\thispagestyle{empty}\fi

\renewcommand{\baselinestretch}{0.965}\small\normalsize

\begin{abstract}
\vspace{-1ex}

\noindent
In contrast to comparing faces via single exemplars,
matching sets of face images increases robustness and discrimination performance.
Recent image set matching approaches typically measure similarities between subspaces or manifolds,
while representing faces in a rigid and holistic manner.
Such representations are easily affected by variations in terms of alignment, illumination, pose and expression.
While local feature based representations are considerably more robust to such variations,
they have received little attention within the image set matching area.
We propose a novel image set matching technique,
comprised of three aspects:
{\bf\emph{(i)}}~robust descriptors of face regions based on local features,
partly inspired by the hierarchy in the human visual system,
{\bf\emph{(ii)}}~use of several subspace and exemplar metrics to compare corresponding face regions,
{\bf\emph{(iii)}}~jointly learning which regions are the most discriminative while finding the optimal mixing weights for combining metrics.
Experiments on LFW, PIE and MOBIO face datasets
show that the proposed algorithm obtains considerably better performance than 
several recent state-of-the-art techniques,
such as Local Principal Angle and the Kernel Affine Hull Method.

\end{abstract}

\vspace{-2ex}
\section{Introduction}
\label{sec:introduction}

A recent trend in image set matching considers image sets as linear subspaces,
with the similarity between the sets derived from the similarity between the subspaces~\cite{Cevikalp_CVPR_2010,Harandi_CVPR_2011,Wolf_KPA_2003,Yamaguchi_1998}.
In almost all subspace based approaches,
faces are represented in a {rigid} and {holistic} manner,
where each face is represented by one feature vector that describes the entire face.
Such a representation implicitly embeds rigid spatial constraints between face components~\cite{Cardinaux_TSP_2006}.

While subspaces are thought of being capable of accommodating the effects of various image variations%
\footnote
  {
  For example, a linear subspace can be used for photometric invariance,
  under the conditions of no shadowing and Lambertian reflectance~\cite{Adini1997_PAMI}.
  }%
,
the magnitude and compounding effect of variations (such as illumination, pose and expression changes)
might overwhelm even the most sophisticated subspace modelling technique.
The relatively poor performance of linear models in such challenging recognition tasks
appears to have roots in the non-linear nature of typical image manifolds \cite{KIM_BOMPA_2007,WANG2008_CVPR,Wolf_KPA_2003}, with much effort directed towards handling the non-linearities (eg.~via kernel extensions~\cite{Harandi_CVPR_2011,Wolf_KPA_2003}\nocite{Harandi_2009} and data clustering~\cite{Fitzgibbon2003_CVPR,Hadid:2009:MLV,WANG2008_CVPR}).

In contrast to rigid face representations,
a face can also be represented by a set of {local features}.
This set can then be processed by a classifier that explicitly allows {relaxed} spatial constraints between face parts.
Such a combination allows for some movement and/or deformations of the face components~\cite{Cardinaux_TSP_2006,Heisele_CVIU_2003,Sanderson_PR_2006},
which in turn leads to a degree of inherent robustness to expression and pose changes~\cite{Heisele_CVIU_2003,Sanderson_PR_2006}
as well as misalignment~\cite{Cardinaux_TSP_2006}.
Examples of such systems include
Elastic Graph Matching~\cite{EBGM_PAMI_1997},
pseudo-2D hidden Markov models~\cite{Cardinaux_TSP_2006},
and ``bag of words'' approaches~\cite{Sanderson_ICB_2009}.

Several studies in the domain of single-image to single-image matching
have shown that non-linear structures can be effectively avoided by local representations.
More precisely, while the structures that describe holistic features tend to be non-linear and complex,
linear structures are good/sufficient tools to approximate local features~\cite{KANADE_CIRA_2003,PENTLAND_CVPR_1994}.
As such, rather than using holistic face representations and relying on a model to handle the resulting non-linear variations,
it might be more appropriate to develop an image set matching technique based on local representations,
while allowing relaxed spatial constraints.

We propose an approach for image set face verification that uses a multitude of local representations and distance metrics,
and employs a learning algorithm to determine which subset of descriptors and their associated metrics
is the most useful for discrimination.
More specifically, from each image two types of robust local descriptors are obtained:
region descriptors and compound region descriptors.
The compound descriptors are inspired by the hierarchical architecture of the human visual system,
where the receptive fields of neurons tend to get larger in order to deal with increasingly complex stimuli~\cite{SERRE:PAMI:2007}.
The descriptors from corresponding regions in two face sets
are pooled and then compared via several distance metrics (instead of relying on only one),
resulting in a high-dimensional similarity vector.
As such, the image set verification problem is converted to a binary problem on similarity features.

\begin{figure}[!t]
  \centering
  \includegraphics[width=0.85\columnwidth]{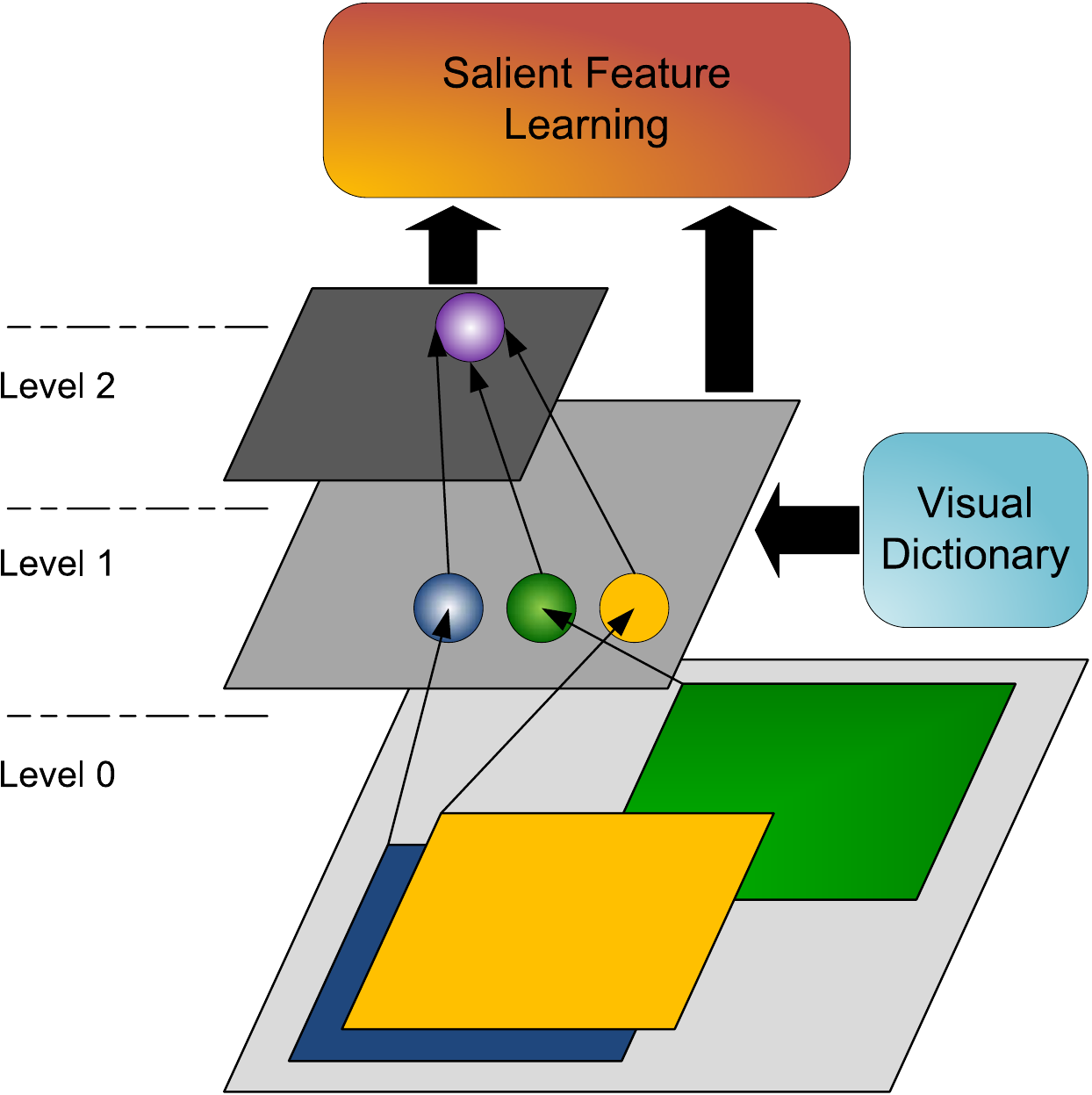}
  \caption
    {
    \small
    Partly inspired by the human visual system~\cite{SERRE:PAMI:2007},
    the proposed approach has a hierarchical structure.
    Level~0 is the image plane.
    Level~1 contains descriptors for regions within the face, with the regions having arbitrary sizes and locations.
    Each descriptor is a probabilistic histogram, obtained using a dictionary of visual words.
    Level~2 contains compound descriptors, generated by aggregating the descriptors from Level~1.
    The descriptors from Level~1 and~2 are fed to a learning mechanism which determines
    which subset of descriptors is the most useful for face verification.
    }
  \label{fig:Hierarchy_Diagram}
\vspace{0.5ex}
\hrule
\end{figure}

By learning to separate similarity vectors representing
matched sets (ie.~sets of the same person)
and mismatched sets (ie.~sets of two persons),
we are in effect jointly determining which regions are the most discriminative while finding the optimal mixing weights for combining metrics. 
Fig.~\ref{fig:Hierarchy_Diagram} shows a conceptual overview of the approach.

We continue the paper as follows.
The feature extraction process is described in Section~\ref{sec:algorithm_DRH}.
The details of the learning approach are given in Section~\ref{sec:learning}.
Comparative evaluations of the proposed method against other image set matching techniques are given in Section~\ref{sec:Experiment}.
The main findings and possible future directions are covered in Section~\ref{sec:conclusions}.

\renewcommand{\baselinestretch}{1.0}\small\normalsize

\vspace{-1ex}
\section{Hierarchical Feature Extraction}
\label{sec:algorithm_DRH}
\vspace{-1ex}

As shown in Fig.~\ref{fig:Hierarchy_Diagram}, the feature extraction is hierarchical in nature, with 3 levels.
The lowest level (level 0), is the image plane.
The details for the feature extraction at levels~1 and~2
are given in Sections~\ref{sec:level_1} and~\ref{sec:level_2}, respectively.

\subsection{Level 1}
\label{sec:level_1}
\vspace{-1ex}

Each descriptor in level 1 corresponds to a relatively large region in the image plane.
The descriptor for region size of \mbox{$p \times p$}, at an arbitrary location, is constructed as follows.
In a similar manner to~\cite{Sanderson_ICB_2009},
the region is split into small overlapping blocks, with each block having a size of {\small $8 \times 8$}.
For each block a histogram of probabilities is calculated,
where each entry in the histogram reflects the similarity of the block to a pre-defined `visual word'.
Each region is represented as the average of all the histograms obtained for the region's blocks.
The procedure is elucidated below.

Each block is represented by a low-dimensional texture descriptor.
For each texture descriptor $\Vec{x}_{r,i}$ obtained from a block in region $r$,
a~probabilistic histogram is computed:

\vspace{-2ex}
\noindent
\begin{small}
\begin{equation}
   \Vec{h}_{r,i}
   \mbox{=}
     \left[
        \frac{ w_{_1} ~ p_{_1}\left(\Vec{x}_{r,i}\right)}{ \sum_{g=1}^{G} w_g p_g\left(\Vec{x}_{r,i}\right) },
       ~\cdots,
       ~\frac{ w_{_G} ~ p_{_G}\left(\Vec{x}_{r,i}\right) }{ \sum_{g=1}^{G} w_g p_g\left(\Vec{x}_{r,i}\right) }
     \right]^T
\label{eqn:prob_histogram}
\end{equation}%
\end{small}%

\noindent
where the $g$-th element in {\small $\Vec{h}_{r,i}$} is the posterior probability of {\small $\Vec{x}_{r,i}$}
according to the $g$-th component of a visual dictionary model.
The visual dictionary model employed here is a convex mixture of Gaussians~\cite{Bishop_2006},
parameterised by {\small $\lambda = \left\{ w_g, \Vec{\mu}_g, \mathbf{C}_g \right\}_{g=1}^{N_G}$},
where {\small $N_G$} is the number of Gaussians,
while {\small $w_g$}, {\small $\Vec{\mu}_g$} and {\small $\mathbf{C}_g$}
are the weight, mean vector and covariance matrix for Gaussian $g$, respectively.
The mean of each Gaussian can be thought of as a particular `visual word'.
The visual dictionary is obtained by pooling a large number of texture descriptors from training images,
followed by employing the Expectation Maximisation algorithm~\cite{Bishop_2006}
to find the dictionary's parameters (\ie, $\lambda$).

In this work we use local texture descriptors based on DCT analysis with illumination normalisation~\cite{Sanderson_ICB_2009}.
However, it is possible to use other texture descriptors,
eg., based on Gabor wavelets~\cite{Lee_PAMI_1996} or Local Binary Patterns~\cite{LBP_PAMI2006}.

Once the histograms are computed for each feature vector from region $r$,
an average histogram for the region is built:
{\small $\Vec{h}_{r,\mathtt{avg}} = \frac{1}{N} \sum\nolimits_{i=1}^{N} \Vec{h}_{r,i}$}.
Due to the averaging operation, in each region there is a loss of spatial relations between face parts.
As such, each region is in effect described by an orderless collection of local features (`bag-of-words').
The loss of spatial relations allows for a degree of misalignment,
pose variations and expression changes~\cite{Cardinaux_TSP_2006,Heisele_CVIU_2003,Sanderson_PR_2006,Sanderson_ICB_2009}.

\subsection{Level 2}
\label{sec:level_2}
\vspace{-1ex}

In the human visual system,
the receptive fields of neurons tend to get larger in order to deal with increasingly complex stimuli~\cite{SERRE:PAMI:2007}.
The responses of complex cells can be pooled from the responses of adjacent simple cells
using `max' or `sum' operations~\cite{HMAX:1999,SERRE:PAMI:2007}.
In a similar manner, we use three configurations for combining the descriptors from level~1, using the `sum' operation.

\begin{figure}[!tb]
\begin{minipage}{\columnwidth}
  \begin{minipage}{0.31\columnwidth}
    \centerline{\includegraphics[width=0.9\textwidth]{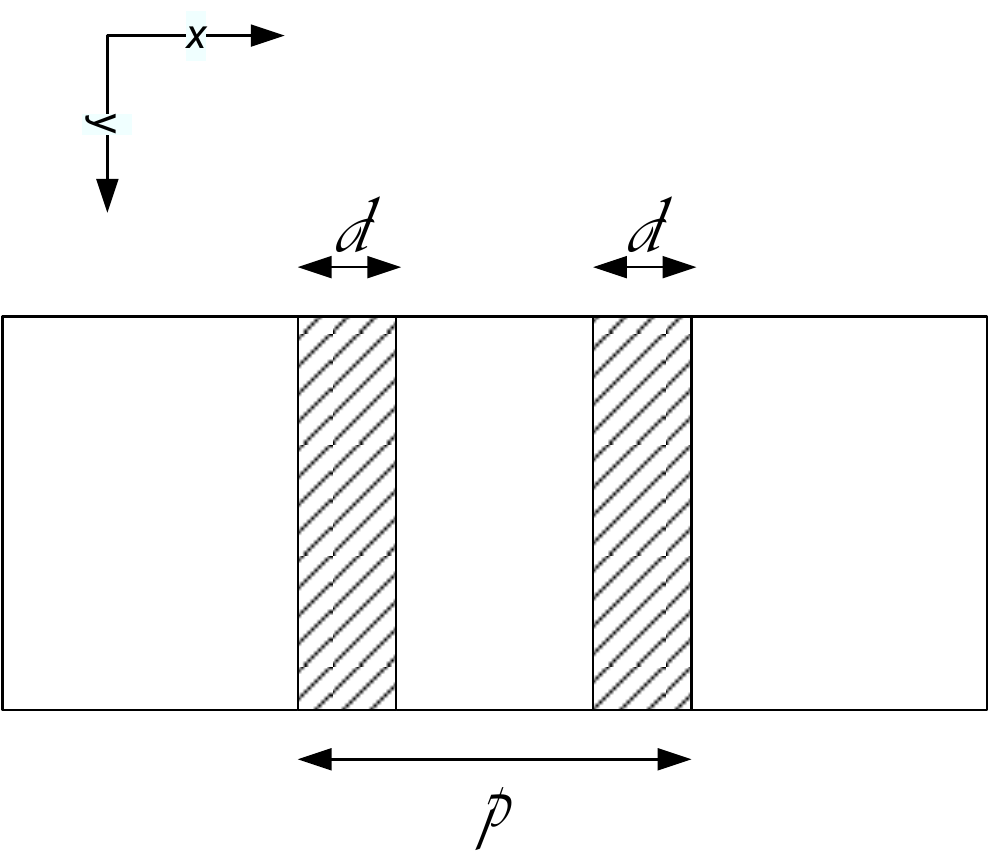}}
  \end{minipage}
  \hfill
  \begin{minipage}{0.31\columnwidth}
    \centerline{\includegraphics[width=0.6\textwidth]{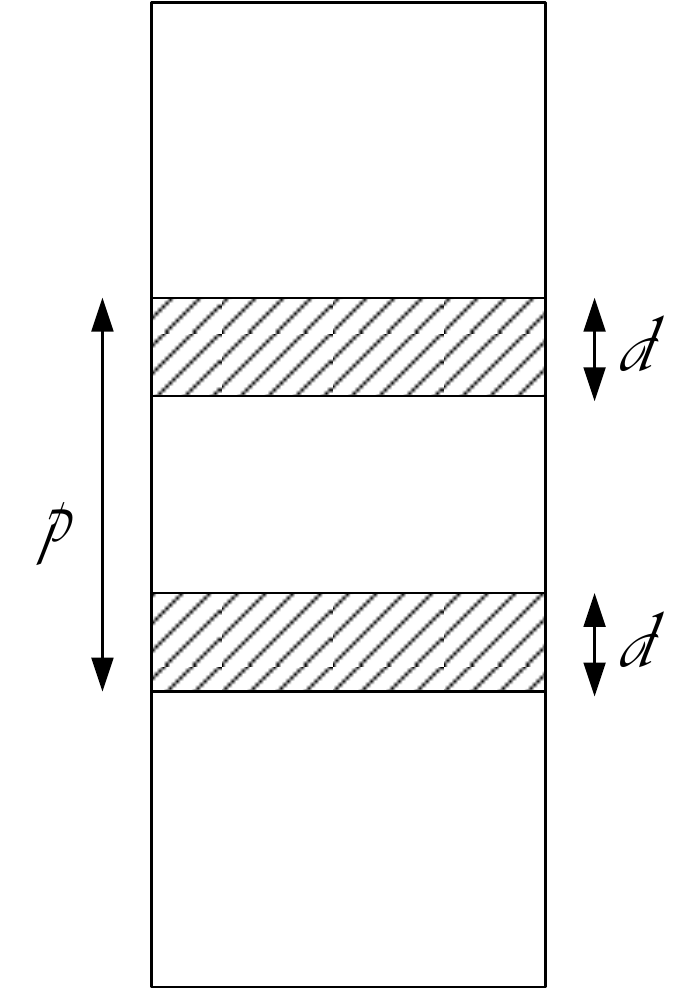}}
  \end{minipage}
  \hfill
  \begin{minipage}{0.31\columnwidth}
    \centerline{\includegraphics[width=0.9\textwidth]{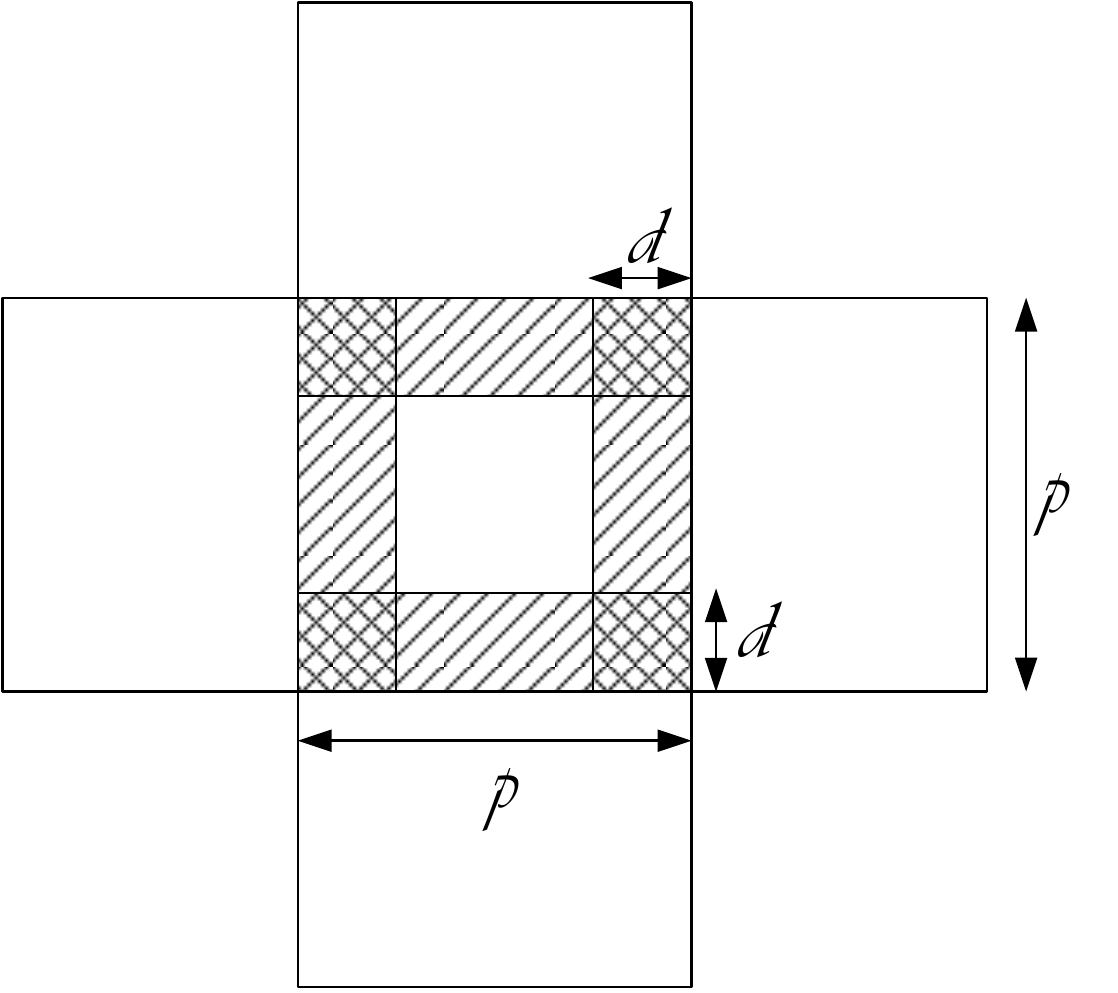}}
  \end{minipage}
\end{minipage}
\begin{minipage}{\columnwidth}
  \begin{minipage}{0.31\columnwidth}
    \centerline{\small\bf (a)}
  \end{minipage}
  \hfill
  \begin{minipage}{0.31\columnwidth}
    \centerline{\small\bf (b)}
  \end{minipage}
  \hfill
  \begin{minipage}{0.31\columnwidth}
    \centerline{\small\bf (c)}
  \end{minipage}
  \caption
    {
    \small
    Compound descriptors are generated by computing the sum over the descriptors from level~1.
    We use three configurations:
    {\bf (a)} for representing horizontally elongated shapes like the eyes and mouth;
    {\bf (b)} for representing vertical elongated shapes such as the nose;
    {\bf (c)} a mixture of~(a) and~(b), for capturing a degree of correlations between shapes such as the nose and mouth.
    }
  \label{fig:Compound_Features}
\end{minipage}
\vspace{-2ex}
\end{figure}

The three configurations are shown in Fig.~\ref{fig:Compound_Features}.
The first configuration is in effect a horizontal shape.
The compound descriptor in this case is a summation of three regions, \ie~simple cells, where the centers of the two outer regions
are located at {\small $(-(p-d),0)$} and {\small $(p-d,0)$} relative to the center of the middle region,
where \mbox{$p \times p$} is the region size.
The second configuration is similar to the first, except a vertical shape is used.
We conjecture the first configuration can be useful for capturing horizontally elongated structures such as the mouth, eyes and eyebrows,
while the second configuration can be useful for representing vertically elongated shapes, such as the nose.

The third configuration is a combination of the previous two shapes and forms a cross shape.
We believe it can be useful for capturing a degree of correlations between the appearance of various face parts.
For example, the shape of the nose might be related to the shape of the mouth.

\vspace{-1ex}
\section{Determining Salient Descriptors}
\label{sec:learning}
\vspace{-1ex}

An image set face verification system needs to determine
whether two sets,  {\small $\mathbb{A}$} and {\small $\mathbb{B}$}, represent the same person.
In general this is accomplished by comparing the similarity between the two sets to a predefined threshold $\tau$.

We assume that the image set $\mathbb{A}$ is comprised of {\small $l$} images.
Each image $i$ is represented by $\nu$ descriptors (histograms from level 1 and 2),
{\small $\Vec{h}_{1}^{[i]}, \Vec{h}_{2}^{[i]}, \cdots, \Vec{h}_{\nu}^{[i]}$},
with each descriptor covering a particular region.
We define a {\it local mode} as a matrix which contains all descriptors for region $j$ from the $l$ images:

\vspace{-1ex}
\begin{small}
\begin{equation}
  \Mat{L}^{\mathbb{A}}_{j} = \left[ \Vec{h}_{j}^{[1]} \mid \Vec{h}_{j}^{[2]} \mid \cdots \mid \Vec{h}_{j}^{[l]} \right],
  ~j = 1, 2, \cdots, \nu
\end{equation}%
\end{small}%

To compare two corresponding local modes from sets {\small $\mathbb{A}$} and {\small $\mathbb{B}$}, 
ie.,~{\small $\Mat{L}_j^{\mathbb{A}}$} and {\small $ \Mat{L}_j^{\mathbb{B}}$},
instead of relying on only one similarity measure,
we propose to use $k$ similarity measures:
{\small $d_{1} \left(\Mat{L}_j^{\mathbb{A}}, \Mat{L}_j^{\mathbb{B}}  \right)$},
{\small $d_{2} \left(\Mat{L}_j^{\mathbb{A}}, \Mat{L}_j^{\mathbb{B}}  \right)$},
{\small $\cdots$},
{\small $d_{k} \left(\Mat{L}_j^{\mathbb{A}}, \Mat{L}_j^{\mathbb{B}}  \right)$}.
We define the overall {\it similarity vector} between sets {\small $\mathbb{A}$} and {\small $\mathbb{B}$}
as containing $k$ similarity measures for each local mode, resulting in a $k\nu$-dimensional vector:

\vspace{-1ex}
\begin{footnotesize}
\begin{equation}
\Vec{S} \left( \mathbb{A},\mathbb{B} \right) = 
\left[
\begin{array}{c}
d_1 ( \Mat{L}_1^{\mathbb{A}} , \Mat{L}_1^{\mathbb{B}} )\\
...\\
d_k ( \Mat{L}_1^{\mathbb{A}} , \Mat{L}_1^{\mathbb{B}} )\\
d_1 ( \Mat{L}_2^{\mathbb{A}} , \Mat{L}_2^{\mathbb{B}} )\\
...\\
d_k ( \Mat{L}_2^{\mathbb{A}} , \Mat{L}_2^{\mathbb{B}} )\\
...\\
d_1 ( \Mat{L}_\nu^{\mathbb{A}} , \Mat{L}_\nu^{\mathbb{B}} )\\
...\\
d_k ( \Mat{L}_\nu^{\mathbb{A}} , \Mat{L}_\nu^{\mathbb{B}} )
\end{array}
\right]
\end{equation}%
\end{footnotesize}

\begin{figure}[!tb]
  \centering
  \includegraphics[width=0.6\columnwidth]{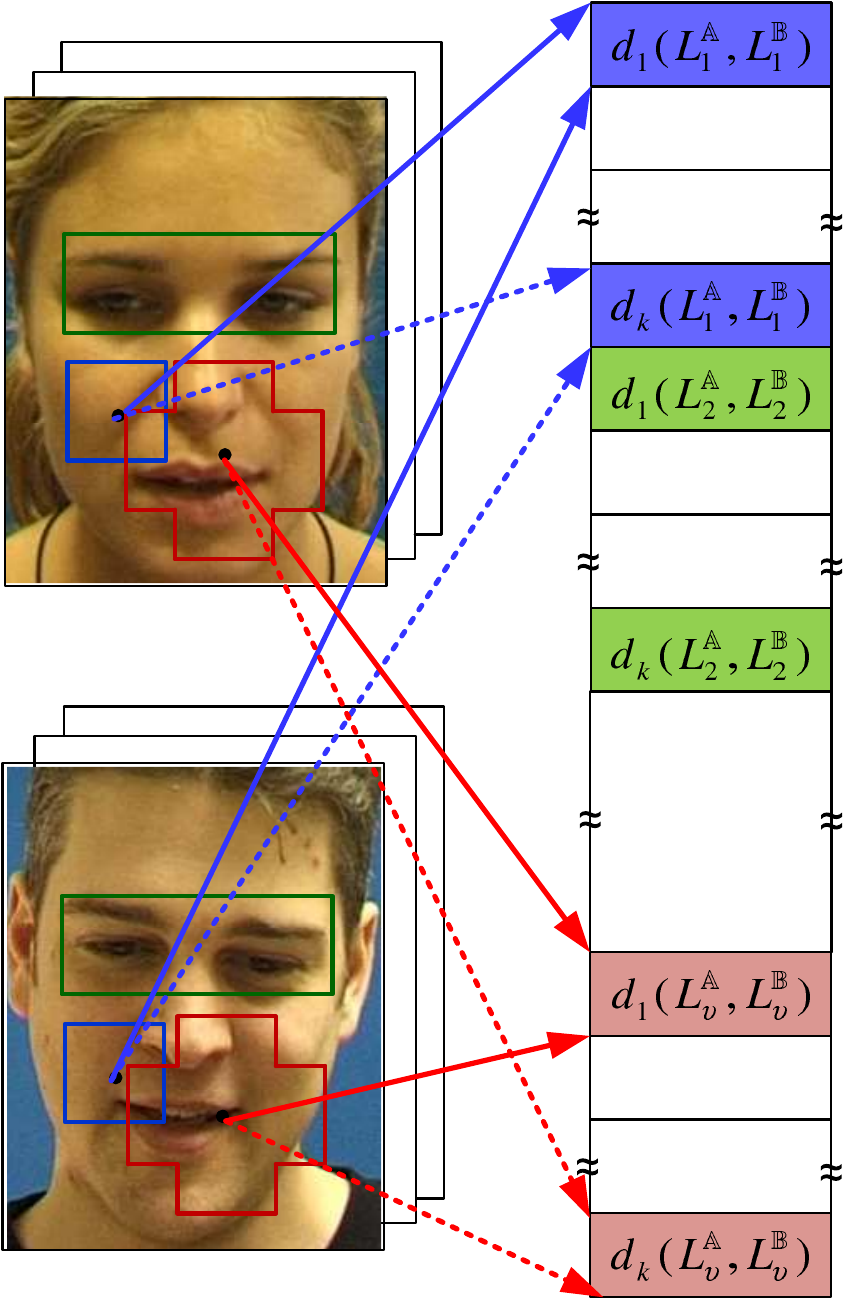}
  \caption
    {
    \small
    Converting the image set verification problem to a binary problem on similarity features.
    Each region in a single image for a given person is described by an average histogram of visual words.
    The corresponding histograms for a particular region across several images form a local mode.
    The corresponding local modes from two people are compared using several distance metrics.
    All the resulting distances for all modes are placed into a similarity vector.
    }
  \label{fig:DistanceFeature}
  \vspace{-2ex}
\end{figure}

\noindent
The image set verification problem is hence converted to a binary classification problem involving similarity vectors.
Figure~\ref{fig:DistanceFeature} provides a graphical interpretation.

We use two families of similarity measures: subspace based, and exemplar based.
For the subspace based measures, we employ the Grassmannian geodesic distance (arc-length) and Binet-Cauchy distance~\cite{HAMM2008_ICML}.
For the exemplar based measures, we use Hausdorff and Modified Hausdorff distances~\cite{MHD_ICPR}.
The two families are elucidated below.

For the subspace based measures,
each local mode {\small $\Mat{L}_i^{\mathbb{A}}$} is modelled by a linear subspace.
A common similarity measure between subspaces is the concept of principal angles~\cite{Yamaguchi_1998}.
If
{\small $\Mat{O}_1 \in \mathbb{R}^{d \times n_1}$}
and
{\small $\Mat{O}_2 \in \mathbb{R}^{d \times n_2}$}
are two linear subspaces in {\small $\mathbb{R}^{d}$}
with minimum rank  {\small $r=\operatorname{min}(rank(\Mat{O}_1,\Mat{O}_2))$},
then there are exactly $r$ uniquely defined principal angles
between {\small $\Mat{O}_1$} and {\small $\Mat{O}_2$}:

\vspace{-1ex}
\noindent
\begin{small}
\begin{equation}
  \cos(\theta_i)
  =
  \max_{\Vec{x}_i \in \Mat{O}_1, ~ \Vec{y}_j \in \Mat{O}_2}
  \Vec{x}_i^T \Vec{y}_j
  \label{eqn:Principal_Angle}
\end{equation}%
\end{small}%

\noindent
subject to \mbox{\small $\Vec{x}_i^T \Vec{x}_i = \Vec{y}_i^T \Vec{y}_i = 1,
\Vec{x}_i^T \Vec{x}_j = \Vec{y}_i^T \Vec{y}_j = 0 , i \neq j$}.
A straightforward method for computing the principal angles is based on Singular Value Decomposition. 
More specifically, the cosines of the principal angles are the singular values of {\small $\Mat{O}_1^T \Mat{O}_2$}:

\vspace{-2ex}
\noindent
\begin{small}
\begin{equation}
  \Mat{O}_1^T \Mat{O}_2
  =
  \Mat{U} \Mat{\Lambda} \Mat{V}^T
  \label{eqn:PA_SVD_1}
\end{equation}%
\end{small}%

\noindent
where the singular values are the diagonal entries of~{\small $\Mat{\Lambda}$}.

Based on the above principal angles, we use two similarity measures:
Grassmannian geodesic distance and Binet-Cauchy distance,
defined respectively as~\cite{HAMM2008_ICML}:

\vspace{-2ex}
\noindent
\begin{small}
\begin{eqnarray}
  d_\mathrm{G} (\Mat{O}_1, \Mat{O}_2) & = & \sum\nolimits_{i} {\theta_i^2} \label{eqn:Geodesic_Distance} \\
  d_\mathrm{BC}(\Mat{O}_1, \Mat{O}_2) & = & \sqrt{1- \prod\nolimits_{i} \cos^2(\theta_i) } \label{eqn:BC_Distance}
\end{eqnarray}%
\end{small}%

For the exemplar based measures, 
local modes are compared using Hausdorff and Modified Hausdorff distances~\cite{MHD_ICPR}.
Given two corresponding local modes {\small $\Mat{L}_i^{\mathbb{A}}$} and {\small $\Mat{L}_i^{\mathbb{B}}$},
the Hausdorff distance (HD) is defined as:

\vspace{-1ex}
\noindent
\begin{small}
\begin{equation}
  d_\mathrm{HD}
  \hspace{-2pt}
  \left(
  \hspace{-2pt}
  \Mat{L}_i^{\mathbb{A}}\hspace{-2pt}, \Mat{L}_i^{\mathbb{B}}
  \hspace{-1pt}
  \right)
  \mbox{=}
  \max
  \left(
    \hspace{-2pt}
    \max_{a \in \mathbb{A}}
    \min_{b \in \mathbb{B}}
    \left \| a \mbox{~-~} b \right \|
    ,
    \max_{b \in \mathbb{B}}
    \min_{a \in \mathbb{A}}
    \left \| a \mbox{~-~} b \right \|
    \hspace{-2pt}
  \right)
  \label{eqn:Hausdorff_Distance}
\end{equation}%
\end{small}%

Intuitively, if the Hausdorff distance is $d$,
then every point of $\mathbb{A}$ must be within a distance $d$ of some point $\mathbb{B}$ and vice versa.
For image processing applications,
Dubuisson \etal~\cite{MHD_ICPR} proposed the modified Hausdorff distance (MHD),
which is more robust against outliers:

\vspace{-1ex}
\noindent
\begin{small}
\begin{equation}
  d_\mathrm{MHD}(\Mat{L}_i^{\mathbb{A}},\Mat{L}_i^{\mathbb{B}})
  =
  \max\left( d_\mathrm{M}(\Mat{L}_i^{\mathbb{A}},\Mat{L}_i^{\mathbb{B}}),d_\mathrm{M}(\Mat{L}_i^{\mathbb{B}},\Mat{L}_i^{\mathbb{A}}) \right)
  \label{eqn:MHD}
\end{equation}%
\end{small}%

\noindent
where
{\small $d_\mathrm{M}(\Mat{L}_i^{\mathbb{A}},\Mat{L}_i^{\mathbb{B}})
  =
  \frac{1}{\left | \mathbb{A} \right |}
  \sum\nolimits_{a \in \mathbb{A}}
  \min_{b \in \mathbb{B}}
  \left \| a-b \right \|
$},
with {\small $\left | \mathbb{A} \right |$} denoting the cardinality of set {\small $\mathbb{A}$}.

Due to the dense nature of the feature extraction process,
a hefty and redundant representation is available for any image,
leading to a very high dimensional similarity vector.
A~further contributing factor to the high dimensionality is the use of four distance metrics per local mode.
As such, instead of blindly feeding the similarity vectors to a standard learning mechanism
such as a Support Vector Machine~\cite{Bishop_2006},
we have elected to use an adapted version of the AdaBoost algorithm~\cite{VIOLA_ADABOOST_2004},
which is more suitable for dealing with such high dimensional problems.

In the adapted AdaBoost, each weak learner works for a single feature each time.
As a result after $Q$ rounds of boosting, $Q$ features are selected.
The adapted version hence has a considerably lower computational complexity than the original version~\cite{ADABOOST_ECOLT_95}:
in a $D$-dimensional problem, $Q$ comparisons are required instead of $Q \times D$.

\vspace{-1ex}
\section{Experiments}
\label{sec:Experiment}
\vspace{-1ex}

In this section we first provide an overview of the image datasets used in the experiments~(Section~\ref{sec:image_datasets}),
followed by a comparative performance evaluation against several benchmark and recent state-of-the-art methods~(Section~\ref{sec:perf_eval}).

\subsection{Image Datasets}
\label{sec:image_datasets}
\vspace{-1ex}

We employed 3 datasets:
Labeled Faces in the Wild (LFW)~\cite{LFW_Database}, CMU PIE~\cite{PIE_Database} and MOBIO~\cite{MOBIO2010ICPR}.
The datasets contain various face orientations, expressions, illumination situations and occlusions.
A verification setup similar to the LFW protocol~\cite{LFW_Database} is used,
where the task is to classify a pair of previously unseen image sets
as either belonging to the same person (matched pair)
or two different persons (mismatched pair).
In all experiments the images are split into three groups:
{\bf (i)}~training,
{\bf (ii)}~development,
{\bf (iii)}~evaluation.
The training group was used purely for constructing the visual dictionary
--- its subjects were never seen in the development and evaluation groups.
Experiments on all datasets were carried out on face images which are closely cropped and downsampled to a size of \mbox{$64 \times 64$}.
Each image set contains three images.
The number of matched pairs and mismatched pairs is the same (balanced),
in order to prevent a bias towards one of the pair types.

For the LFW dataset, 620 pairs of image sets were generated,
with 310 pairs for development group and 310 pairs for evaluation group.
The generic subset from LFW view 1 was used for the training group.

For the CMU PIE dataset, we used the near frontal poses (C05, C07, C09, C27 and C29),
resulting in 170 images per subject with various illuminations and expressions.
We randomly selected 8 subjects for the training group while development and evaluation groups each have 30 subjects.
1,200 pairs of images were generated, with the development and evaluation groups having 600 pairs each.

The MOBIO dataset contains images captured from mobile devices.
The quality of the images is generally poor with blurring from motion and smudged lenses,
as well as changes in illumination between scenes.
A Haar-based cascade classifier~\cite{VIOLA_FD_2004} was used to locate faces in each frame.
The eyes within each face are located using a similar cascade classifier.
If no eyes are located, their approximate location is inferred from the size of the face bounding box.
The faces are then resized and cropped such that the eyes are centered with a 32-pixel inter-eye distance.
We used the development subset of MOBIO,
which contains 1,500 probe videos from 20 females and 27 males.
We generated 832 pairs of images for the development group and 800 pairs for the evaluation group.
The background data subset was used as the training group.

\subsection{Comparative Performance Evaluation}
\label{sec:perf_eval}
\vspace{-1ex}

The proposed approach is compared against
several benchmark methods as well as recent state-of-the-art methods.
The evaluated methods are representative techniques for exemplar-based and subspace-based approaches.

The exemplar-based techniques are:
Laplacianface~\cite{LAPLACIANFACE_2005}, Local Binary Pattern (LBP)~\cite{LBP_PAMI2006}, Multi-Region Histograms (MRH)~\cite{Sanderson_ICB_2009}\nocite{Armadillo_2010}, and Local Facial Features (LFF)~\cite{KANG_EURASIP_2010}. The subspace-based techniques are: Mutual Subspace Method (MSM)~\cite{Yamaguchi_1998}, Kernel Affine Hull Method (KAHM)~\cite{Cevikalp_CVPR_2010}, and Local Principal Angle (Local-PA)~\cite{LI_ACCV_2009}.

We note that the above approaches can also be classified as either local or holistic
in terms of the underlying feature extraction.
LBP, MRH, LFF, Local-PA and the proposed approach are in the local based category,
while Laplacianface, MSM and KAHM are in the holistic based category.

Similarity judgements in exemplar-based methods were carried out using the Modified Hausdorff Distance~(MHD)~\cite{MHD_ICPR}.
The KAHM approach used a linear kernel with the parameters tuned according to the recommendations made in~\cite{Cevikalp_CVPR_2010}.
The best results are reported.
For LBP, uniform histograms with \mbox{$(8,1)$} neighbourhoods are employed.
The LBP block size was selected empirically as \mbox{$7 \times 9$}.
In
Laplacianface,
the subspace dimensions were set by retaining enough leading eigenvectors to account for $98\%$
of the overall energy in the eigen-decomposition.
In Local-PA, the block size was \mbox{$16 \times 16$}, also obtained empirically.

Based on preliminary experiments,
the proposed approach used the following parameters:
the size of each region is $24 \times 24$,
dimension of each DCT-based texture descriptor is $15$,
and the number of visual words in the dictionary is 1024.

To generate compound descriptors,
the distance between centers of simple cells (regions in the image plane) was selected as 4, 8 and 12.
For images of size \mbox{$64 \times 64$}, this results in 1681 direct regions and 8153 compound regions.
As four distance metrics are used for each local mode,
the dimensionality of the resulting similarity vector for each image set pair is 39336.
The discrimination performance appears to stabilise with a subset of 150 similarity features,
as selected by the AdaBoost algorithm.

\begin{figure}[!b]
\vspace{-3ex}
  \begin{minipage}{1\columnwidth}
    \begin{minipage}{\textwidth}
      \centering
      \begin{minipage}{0.1\textwidth}
        ~
      \end{minipage}
      \begin{minipage}{0.28\textwidth}
        \centering
        \includegraphics[width=\textwidth,keepaspectratio]{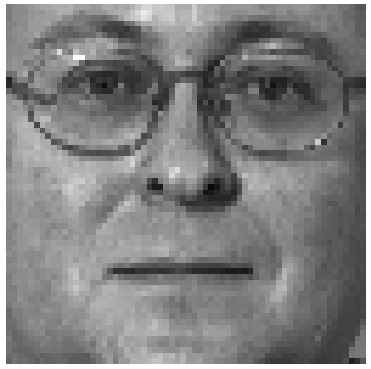}
      \end{minipage}
      \begin{minipage}{0.1\textwidth}
        ~
      \end{minipage}
      \begin{minipage}{0.28\textwidth}
        \centering
        \includegraphics[width=\textwidth,keepaspectratio]{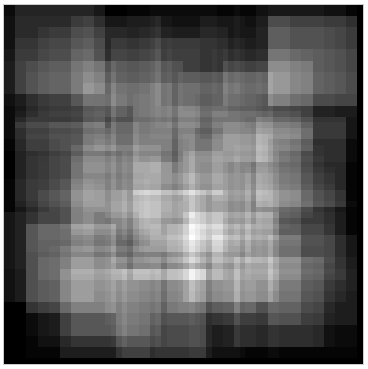}
      \end{minipage}
      \begin{minipage}{0.1\textwidth}
        ~
      \end{minipage}
    \end{minipage}
    \begin{minipage}{\textwidth}
      \centering
      \begin{minipage}{0.1\textwidth}
        ~
      \end{minipage}
      \begin{minipage}{0.28\textwidth}
        \centering
        {\bf\small (a)}
      \end{minipage}
      \begin{minipage}{0.1\textwidth}
        ~
      \end{minipage}
      \begin{minipage}{0.28\textwidth}
        \centering
        {\bf\small (b)}
      \end{minipage}
      \begin{minipage}{0.1\textwidth}
        ~
      \end{minipage}
    \end{minipage}
  \end{minipage}
  \caption
    {
    \small
    An example of the cumulative weights for face regions selected by the boosting algorithm:
    {\bf (a)}~cropped face from PIE;
    {\bf (b)}~brighter regions correspond to higher cumulative weights.
    }
  \label{fig:Selected_Patches}
\end{figure}

An example of cumulative weights of the most discriminant local modes obtained by the boosting algorithm
is shown in Fig.~\ref{fig:Selected_Patches}.
Cumulative weight for a pixel {\small $I(x,y)$} is defined as the sum of the weights of the selected regions that include the pixel.
Most of regions are selected from the inner part of the face,
with stress on the regions around the mouth, nose and eyes.

The comparative results are shown in Table~\ref{tab:Comparison_results},
with the verification accuracy defined as the average of the accuracy on matched and mismatched pairs.
The relatively poor performance of the Laplacianface approach implies the difficulty of the recognition task,
considering that the method is expected to perform relatively well if the imaging conditions
do not differ greatly between training and test datasets.

The results show that in all experiments local approaches prevail over holistic techniques.
This confirms the premise of this work:
{\it relaxed local representations are more \mbox{robust} than rigid holistic representations}.
Among exemplar-based methods, MRH and LFF outperform Laplacianface and LBP.
Among the subspace approaches, Local~PA outperforms MSM.
We note that KAHM is marginally superior to MSM (with the exception of CMU-PIE),
however LBP+KAHM significantly outperforms MSM for all experiments.
This is consistent with the results reported in~\cite{Cevikalp_CVPR_2010}.

The proposed approach surpasses all other methods by a considerable margin on the LFW and PIE datasets.
On LFW, the performance difference to LFF, the nearest competing approach, is 7.7 percentage points.
On PIE, the improvement over the nearest method is close to 13 percentage points.

\begin{table}[!tb]
  \centering
  \caption
    {
    \small
    Average verification accuracy on LFW, PIE and MOBIO datasets.
    The methods are grouped into two categories:
    {\bf (a)}~exemplar based,
    and
    {\bf (b)}~subspace based.
    The proposed method uses both exemplar and subspace based similarity metrics.
    }
  \label{tab:Comparison_results}
  \vspace{0.5ex}
  \begin{small}
  \begin{tabular}{l l c c c c}
    \toprule
    \hspace{-1ex}& {\bf Method}                                  &{\bf LFW}~  &~{\bf PIE}~   &\hspace{-1ex}{\bf MOBIO}\hspace{-1ex}   & overall\\
    \toprule
    \hspace{-1ex}\multirow{4}{*}{\footnotesize (a) $\left\{ \begin{array}{l} \\ \\ \\ \\ \end{array} \right.$}\hspace{-5ex}
    & Laplacian~\cite{LAPLACIANFACE_2005} + MHD\hspace{-2ex}     & 65.48       & 69.17        & 85.50        & 73.38 \\
    & LBP~\cite{LBP_PAMI2006} + MHD                 & 79.35       & 78.17        & 94.75        & 84.09 \\
    & MRH~\cite{Sanderson_ICB_2009} + MHD           & 86.45       & 75.50        & 96.75        & 86.23 \\
    & LFF~\cite{KANG_EURASIP_2010}                  & 88.06       & 78.17        & 97.75        & 87.99 \\
    \midrule
    \hspace{-1ex}\multirow{4}{*}{\footnotesize (b) $\left\{ \begin{array}{l} \\ \\ \\ \\ \end{array} \right.$}\hspace{-5ex}
    & MSM~\cite{Yamaguchi_1998}                     & 65.48       & 71.33        & 90.13        & 75.65 \\
    & Local-PA~\cite{LI_ACCV_2009}                  & 67.10       & 77.17        & 92.50        & 78.92 \\
    & KAHM~\cite{Cevikalp_CVPR_2010}                & 66.13       & 67.83        & 90.38        & 74.78 \\
    & LBP + KAHM~\cite{Cevikalp_CVPR_2010}          & 73.22       & 76.00        & 95.38        & 81.53 \\
    \midrule
    & {\bf Proposed method}                         & {\bf 95.80} & {\bf 91.00}  & {\bf 100.00} & {\bf 95.60} \\
    \bottomrule
  \end{tabular}%
  \end{small}%
  \vspace{-2ex}
\end{table}

\section{Main Findings and Future Directions}
\label{sec:conclusions}
\vspace{-1ex}

We have proposed a novel image set matching technique for face verification,
comprised of three aspects:
{\bf (i)} robust descriptors of face regions based on local features, partly inspired by the hierarchy in the human visual system,
{\bf (ii)} use of several subspace and exemplar metrics to compare corresponding face regions,
{\bf (iii)} jointly learning which regions are the most discriminative while finding the optimal mixing weights for combining metrics.
Experiments on LFW, PIE and MOBIO face datasets show that the proposed algorithm obtains considerably better performance
than several recent state-of-the-art techniques, such as Local Principal Angle and the Kernel Affine Hull Method.

We note that the region descriptors used in Section~\ref{sec:algorithm_DRH} somewhat resemble Sparse Representation (SR) and dictionary learning,
as they are obtained through an over-complete visual dictionary~\cite{ELAD_SR_BOOK_2010}.
While SR methods usually utilise greedy algorithms like Matching Pursuit or convex optimisation~\cite{ELAD_SR_BOOK_2010}
(which are computationally expensive),
the descriptors here are obtained through closed-form equations.
This is useful in large-scale data processing applications.

While the learning method presented here is specific to a verification system (ie.~binary classification),
extension to arbitrary {\small $M$}-class discrimination problems is possible.
An {\small $M$}-class problem can be converted into a binary problem
via the use of intra- and inter-personal spaces~\cite{Moghaddam2000_PR}.
More specifically, instead of characterising class clusters,
it is possible to characterise what kind of image variation is typical for the same person and what is for different persons.
Theoretically this is achieved by training a binary classifier on the differences between two samples,
ie.~{\small $\Delta = S_1-S_2$}.
Based on learning the differences, two samples (here two sets obtained from a specific local region)
are considered as representing the same person if they are classified as intra-personal variation.
Conversely, two samples represent two unique individuals if their difference is classified as extra-personal variation.

\section*{Acknowledgements}

NICTA is funded by the Australian Government as represented by the {\it Department of Broadband, Communications and the Digital Economy},
as well as the Australian Research Council through the {\it ICT Centre of Excellence} program.

\balance
{\footnotesize
\bibliographystyle{ieee}
\bibliography{references}
}

\end{document}